\documentclass{article}
\usepackage{spconf,amsmath,graphicx}
\usepackage{algorithm}
\usepackage{algorithmic}
\usepackage{color}
\usepackage{subfigure}



\def\DNNt{DNN$_{\text{2048}}$}
\def\DNNf{DNN$_{\text{4096}}$}
\def\SCAb{SCA$_{\text{B}}$}
\def\SCAa{SCA$_{\text{A}}$}
\def\WERb{WER$_{\text{B}}$}
\def\WERa{WER$_{\text{A}}$}
\def\GLASSOo{gLasso$_{\text{O}}$} 
\def\GLASSOi{gLasso$_{\text{I}}$} 
\def\Lo{L2$_{\text{O}}$}
\def\Li{L2$_{\text{I}}$} 

\title{AUTOMATIC NODE SELECTION FOR DEEP NEURAL NETWORKS \\USING GROUP LASSO REGULARIZATION}
\name{{\em Tsubasa Ochiai$^{1,\ast}$, Shigeki Matsuda$^{1}$, Hideyuki Watanabe$^{2}$, and Shigeru Katagiri$^{1}$}}
\address{
$^{1\hspace{0.5mm}}$Graduate School of Science and Engineering, Doshisha University, Kyoto, Japan \\
$^{2\hspace{0.5mm}}$Advanced Telecommunications Research Institute International, Kyoto, Japan \\
$^{\ast\hspace{0.5mm}}$eup1105@mail4.doshisha.ac.jp
\thanks{This work was supported in part by JSPS Grants-in-Aid for Scientific Research No. 26280063, MEXT-Supported Program
``Driver-in-the-Loop,'' and Grant-in-Aid for JSPS Fellows.}}

\begin{document}
\ninept
\maketitle
\begin{abstract}
%\vspace{-2.0mm}
We examine the effect of the Group Lasso (gLasso) regularizer in selecting the salient nodes of Deep Neural Network (DNN) hidden layers by applying a DNN-HMM hybrid speech recognizer to TED Talks speech data. We test two types of gLasso regularization, one for outgoing weight vectors and another for incoming weight vectors, as well as two sizes of DNNs: 2048 hidden layer nodes and 4096 nodes. Furthermore, we compare gLasso and L2 regularizers. Our experiment results demonstrate that our DNN training, in which the gLasso regularizer was embedded, successfully selected the hidden layer nodes that are necessary and sufficient for achieving high classification power. 
\end{abstract}

\section{INTRODUCTION}
\label{sec:introduction}
\vspace{-0.0mm}
Motivated by the recent rise of Deep Neural Networks (DNNs), we studied their utility for speech recognition by developing DNN-HMM hybrid speech recognizers \cite{ochiai2014,ochiai2015}. A DNN's high capability is mainly derived from a deep layer structure and a massive amount of network weight parameters. Therefore, roughly speaking, the larger and deeper the network, the higher its performances are. However, a large network is obviously undesirable from the viewpoints of computational load and memory size; it is also unfavorable from the viewpoint of controlling training robustness to unseen data. To meet this requirement for finding a small, necessary, and sufficient DNN structure, several approaches have reshaped the network structure \cite{xue2013,xue2014,ochiai2016} or pruned the network nodes \cite{tianxing2014}. However, these methods assumed retraining or adapting a size-reduced network for high discriminative power. Therefore, a need obviously exists for developing training methods that can automatically (without additional retraining and adaptation) find a small but sufficient DNN structure. 
 
Seeking such development, we propose in this paper a new training scheme that embeds group Lasso (gLasso) regularization \cite{yuan2006,simon2013} for hidden layer weight vectors. To examine its feasibility, we implement it in a large-scale DNN-HMM speech recognizer and conduct various evaluation experiments in a difficult task: TED Talks. 

Quite recently, a gLasso-based idea similar to ours was reported \cite{scardapane2016}.
However, we independently studied it and elaborated the characteristics of its scheme in large-scale experiments.
In Section 4, we will discuss the relationship between our work and related works, including \cite{scardapane2016}.

%----------------%
\section{DEEP NEURAL NETWORK TRAINING ASSOCIATED WITH NODE SELECTION}
\label{sec:Automatic}
\subsection{Computational Procedures in DNNs}
\label{sec:procedureDNN}

We consider a standard $L$-layer DNN, where the $0$th layer is an input layer, the $L$-th layer is an output layer, and the $1$st through $(L-1)$-th layers are hidden layers. In this network, the $l$-th layer consists of $N_{l}$ nodes, each of which is associated with a bias coefficient. The nodes are also fully connected among the adjacent layers, and each connection is associated with a weight parameter.  

When a vector pattern is given to the input layer, the network feeds the data from the input layer to the output layer in the following layer-by-layer manner: 
\vspace{-1.0mm}
\begin{align}
\mathbf{a}^{l} = \mathbf{W}^{l} \mathbf{z}^{l-1} + \mathbf{b}^{l}, \text{  and  }
\mathbf{z}^{l} = \sigma ( \mathbf{a}^{l} ),
\vspace{-2.0mm}
\label{eq:activationVector}
\end{align}
where $\mathbf{W}^{l}$ is the weight matrix between the $(l-1)$-th and $l$-th layers, $\mathbf{b}^{l}$ is the bias vector of the $l$-th layer, $\mathbf{z}^{l-1}$ is the output vector from the $(l-1)$-th layer, $\mathbf{a}^{l}$ is the activation vector in the $l$-th layer, and $\sigma( \, )$ is an activation function. Here, the $i$-th row and $j$-th column element of $\mathbf{W}^{l}$ are $w^{l}_{ij}$, which is the weight parameter between the $j$-th node of the $(l-1)$-th layer and the $i$-th node of the $l$-th layer; the vectors are detailed as follows: $\mathbf{a}^{l}=[a^{l}_{1} \cdots a^{l}_{i} \cdots a^{l}_{N_{l}}]$, $\mathbf{z}^{l-1}=[z^{l-1}_{1} \cdots z^{l-1}_{j} \cdots z^{l-1}_{N_{l-1}}]$ and $\mathbf{b}^{l}=[b^{l}_{1} \cdots b^{l}_{i} \cdots b^{l}_{N_{l}}]$, where $a^{l}_{i}$, $z^{l-1}_{j}$, and $b^{l}_{i}$ are respectively the aggregated input to the $i$-th node of the $l$-th layer, the output from the $j$-th node of the $(l-1)$-th layer, and the bias of the $i$-th node of the $l$-th layer. 
%
%----------------%
\begin{figure}[t]
%
%\begin{minipage}[b]{1.0\linewidth}
%  \centering
%  \centerline{\includegraphics[width=4.5cm]{network}}
%  \vspace{-0.0mm}
%\end{minipage}
\begin{minipage}[b]{1.0\linewidth}
	\centering
\subfigure[Outgoing]{
\includegraphics[width=2.1cm]{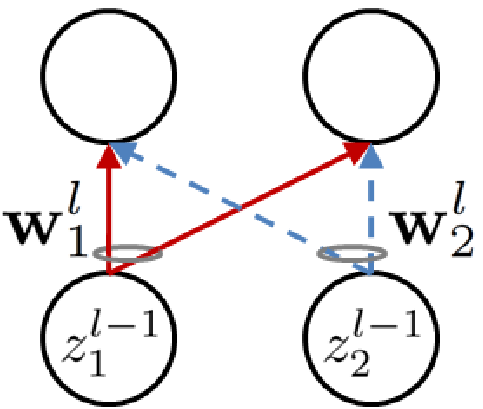}
}
\hspace{1.0cm}
\subfigure[Incoming]{
\includegraphics[width=2.1cm]{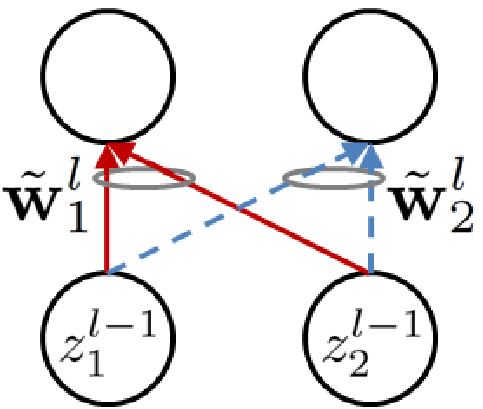}
}
\end{minipage}
\vspace{-5.0mm}
\caption{Schematic explanation of outgoing/incoming weight vector grouping.}
\label{fig:network}
\vspace{-5.0mm}
\end{figure}
%----------------%
%
%----------------%
\subsection{Node Selection without Activation Changes}
\label{sec:nodeSelectionPrinciple}

Figure \ref{fig:network} illustrates an example situation of the node connections between the $(l-1)$-th layer (of the $2$ nodes) and $l$-th layer (of the $2$ nodes). As shown on its left and right sides, we can basically group the connections in two ways: 1) \textit{outgoing} bundle or 2) \textit{incoming} bundle. Based on these groupings, we further group their corresponding weights in outgoing weight vectors ($\mathbf{w}^{l}_{1}$ and $\mathbf{w}^{l}_{2}$) or incoming weight vectors ($\tilde{\mathbf{w}}^{l}_{1}$ and $\tilde{\mathbf{w}}^{l}_{2}$). 
Here, the outgoing weight vectors are column vectors, which form the  columns of weight matrix $\mathbf{W}^{l}$; the incoming weight vectors form the rows of $\mathbf{W}^{l}$.  

In the figure, for example, if the norm of $\mathbf{w}^{l}_{2}$ is nearly zero and the norm for $\mathbf{w}^{l}_{1}$ exceeds zero, $\mathbf{a}^{l}$ is essentially based only on $z^{l-1}_{1}$, i.e., the outputs from the $1$st (${\nu}^{l-1}_{1}$) node in the $(l-1)$-th layer. In this situation, even if we remove node ${\nu}^{l-1}_{2}$, $\mathbf{a}^{l}$ has no large changes and retains the DNN performances that were gained before node pruning. 
On the other hand, again for example, if the norm of $\tilde{\mathbf{w}}^{l}_{2}$ is nearly zero and the norm for $\tilde{\mathbf{w}}^{l}_{1}$ exceeds zero, $\mathbf{a}^{(l+1)}$ is essentially based on $z^{l}_{1}$ and $b^{l}_{2}$ ($\approx z^{l}_{2}$). In this situation, even if we prune node ${\nu}^{l}_{2}$, $\mathbf{a}^{(l+1)}$ has no large changes and the DNN performances accordingly remain almost the same. 
However, because the node is pruned based on its weight vector norm, its bias remains as a constant.
Therefore, we have to remember here to shift the bias for the pruned node to the upper layer nodes. 

As above, regardless of the weight vector grouping, if we can control the values of the weight vector norm in the DNN training, or more precisely, selectively minimize the norm values, we can also achieve a mechanism that automatically selects nodes without changes in the performances of the trained DNNs. 

%----------------%
\subsection{DNN Training with gLasso and L2 Regularizers}
\label{sec:DNNtraining}
\subsubsection{Definition of regularized loss}
The node selection in \ref{sec:nodeSelectionPrinciple} is naturally expected to be done so that only the salient nodes that produce useful outputs for increasing the DNN's performances are retained. If DNN training is done to concurrently minimize a loss function that reflects the DNN's classification errors and a function that reflects the norm values of the outgoing/incoming weight vectors, the unnecessary nodes, which produce useless outputs (for the DNN performances), are automatically disabled by their corresponding small norm values and only the salient nodes will be retained. 

To implement such training, we adopt gLasso regularization for the outgoing or incoming weight vectors in standard DNN training using Cross Entropy (CE) loss. Focusing on the outgoing weight vector case, we formalize our proposed training scheme below. The resulting formalization is basically common to that in the incoming weight vector case. Note that the target layer for grouping is different between types of grouping. 

Our training scheme starts by defining regularized loss $L( \, )$ as follows: 
%
%\vspace{-1.0mm}
%\begin{equation}
%L( \mathbf{\Lambda} ) = E( \mathbf{\Lambda} ) + \alpha \hat{R}_{\text{group}}( \{ \mathbf{W}^{l} \}^{L}_{l=2} ) + \beta \hat{R}_{\text{L2}}( %\mathbf{W}^{1}, \mathbf{b}^{l} ),
%\label{eq:regularizedLoss}
%\end{equation}
\begin{equation}
L( \mathbf{\Lambda} ) = E( \mathbf{\Lambda} ) + \alpha \hat{R}_{\text{group}}( \{ \mathbf{W}^{l} \}^{L}_{l=2} ) + \beta \hat{R}_{\text{L2}}( \mathbf{W}^{1}, \{ \mathbf{b}^{l} \}^{L}_{l=1} ),
\label{eq:regularizedLoss}
\end{equation}
%
%\vspace{-1.0mm}
where $\mathbf{\Lambda}$ is a set of trainable parameters of DNN, i.e., $\mathbf{\Lambda} = \{ \{ \mathbf{W}^{l} , \mathbf{b}^{l} \}^{L}_{l=1} \}$, $E( \, )$ is the CE loss, $\hat{R}_{\text{group}}( \, )$ $\hat{R}_{\text{L2}}( \, )$ are respectively the gLasso and L2 regularizers, and $\alpha$ and $\beta$ are positive constants that control the effects of their corresponding regularizers. Here the gLasso regularizer is specified as 
%
%\vspace{-1.0mm}
\begin{align}
\hat{R}_{\text{group}}( \{ \mathbf{W}^{l} \}^{L}_{l=2} ) 
= \sum_{l=2}^{L} R_{\text{group}}( \mathbf{W}^{l} ) , 
\end{align}
%
%\vspace{-1.0mm}
where 
%
%\begin{align}
$R_{\text{group}}( \mathbf{W}^{l} ) 
= \sum_{j=1}^{N_{l-1}} \| \mathbf{w}^{l}_{j} \|$ , 
%\end{align}
%
%\vspace{-1.0mm}
and the L2 regularizer is specified as
%
%\vspace{-1.0mm}
%\begin{align}
%\hat{R}_{\text{L2}}( \mathbf{W}^{1}, \mathbf{b}^{l} ) 
%= R_{\text{L2}}( \mathbf{W}^{1} ) + \sum_{l=1}^{L} R_{\text{L2}}( \mathbf{b}^{l} ) , 
%\end{align}
\begin{align}
\hat{R}_{\text{L2}}( \mathbf{W}^{1}, \{ \mathbf{b}^{l} \}^{L}_{l=1} ) 
= R_{\text{L2}}( \mathbf{W}^{1} ) + \sum_{l=1}^{L} R_{\text{L2}}( \mathbf{b}^{l} ) , 
\end{align}
%
%\vspace{-1.0mm}
where 
%
%\vspace{-1.0mm}
%\begin{align}
$R_{\text{L2}}( \mathbf{W}^{l} ) = \frac{1}{2} \| \mathbf{W}^{l} \|^{2} $ and 
$R_{\text{L2}}( \mathbf{b}^{l} ) = \frac{1}{2} \| \mathbf{b}^{l} \|^{2}$ .
%\end{align}
%

%----------------%
\subsubsection{Effects of gLasso regularization}
The status of $\mathbf{\Lambda}$ that corresponds to the minimum of $L( \mathbf{\Lambda} )$ will provides high classification power with selected network nodes. To find that status, we minimize $L( \mathbf{\Lambda} )$ using the standard gradient-based optimization procedure. 

In the above gradient-based loss minimization for $L( \mathbf{\Lambda} )$, the gradient terms of the regularizers play a key role in accelerating the minimization. In the L2 regularizer case, the gradient of $R_{\text{L2}}( \, )$ is given as follows, for example, in terms of $\mathbf{w}^{l}_{j}$: 
\begin{equation}
\frac{\partial R_{\text{L2}}( \mathbf{W}^{l} )}{\partial \mathbf{w}^{l}_{j}} = \mathbf{w}^{l}_{j}. 
\label{eq:gradientL2}
\end{equation}
In contrast, the gradient of the gLasso regularizer becomes  
\begin{equation}
\frac{\partial R_{\text{group}}( \mathbf{W}^{l} )}{\partial \mathbf{w}^{l}_{j}} = \frac{\mathbf{w}^{l}_{j}}{ \| \mathbf{w}^{l}_{j} \|} . 
\label{eq:gradientGLASSO}
\end{equation}
Eq. (\ref{eq:gradientGLASSO}) clearly indicates the effects of gLasso regularization; the (``outgoing'' in this case) weight vector rapidly converges to zero when its norm is small. As a result, the node outputs, which are less useful for the minimization of $L( \, )$, are disabled in an early stage of the loss minimization, and the retained node outputs will be dominantly used for the loss minimization.   

%----------------%
\subsection{Node Selection Procedure}
If the weight vector norm becomes sufficiently small after the above loss minimization, we can simply prune its corresponding nodes using a threshold value ($\theta$). The node selection procedure is summarized as follows: 1) calculate the {\GLASSOo} norm (column norm) or the {\GLASSOi} norm (row norm) for all of the hidden layer nodes, 2) prune the nodes whose norm values are less than $\theta$, and 3) shift the bias for the pruned node to the upper layer nodes, especially in the {\GLASSOi} case. 
(See \ref{sec:settings} for the definitions of {\GLASSOo} and {\GLASSOi}.)

%We summarize this node selection (pruning) procedure in \textbf{Algorithm} ``Node selection procedure''.  

%\begin{algorithm}                      
%\caption{{\footnotesize Node selection procedure}}         
%\label{alg}                          
%\begin{algorithmic}                  
%\FOR{$l=2$ to $L$}
%\FOR{$i=1$ to $N_{l-1}$}
%\IF{$\| \mathbf{w}^{l}_{i} \|_{2} < \theta$}
%\STATE {\footnotesize prune $i$-th node at $l$-th layer}
%\ENDIF
%\ENDFOR
%\ENDFOR
%\end{algorithmic}
%\end{algorithm}

%----------------% 
\section{Experiments}
\label{sec:experiments}
\subsection{Settings}

To evaluate our proposed DNN training and node selection method, we conducted speech recognition experiments using our DNN-HMM hybrid speech recognizer %(see \ref{sec:hybridSpeechRecognizer} for details) 
over a TED Talks corpus that consists of lecture speech data spoken by $793$ speakers. For the evaluation experiments, we divided the data into the following three sets: training data by $774$ speakers, validation data by $8$ speakers, and testing data by $11$ speakers. The total length of the training data was about $62$ hours, the number of senone classes was 3,944, and the vocabulary size was $179,604$.  

We represented the input speech as a series of acoustic feature vectors, each of which was based on Mel-scale Frequency Cepstrum Coefficients and of 429 dimensions. See \cite{ochiai2014} for the details of the input pattern definitions. 
 
In our DNN-HMM hybrid speech recognizer, the front-end DNN part was the $6$-layer Multi-Layer Perceptron ($L=6$ in \ref{sec:procedureDNN}) with a sigmoid activation function, and the post-end HMM part used a context-dependent acoustic model and a 3-gram language model. 
We initialized the DNN part by the RBM pre-training \cite{hinton2006}. 
Independently of the DNN part, we developed a HMM part using Boosted MMI training \cite{povey2008}.

The main target of our DNN training was to increase the senone classification accuracy (SCA), whose categories were indicated by the DNN outputs. However, we also evaluated the training results using the word error rate (WER) obtained by the post-end HMM part of the hybrid recognizer. 

%----------------%
\subsection{Evaluation procedures} 
\label{sec:settings}
We ran gradient-based training for minimizing $L( \, )$ in (\ref{eq:regularizedLoss}) several times, along with changing such hyper-parameters as $\alpha$ and $\beta$ in (\ref{eq:regularizedLoss}). 
Nevertheless, for simplicity, we restricted $\beta$ to $\beta = 0.1 \alpha$ or $\beta = 0$. 
In every training run, we repeated twenty epochs, in each of which the whole set of training samples was used only once. 
Among the different trained statuses of $\mathbf{\Lambda}$, we selected the one that produced the highest SCA value over the validation data, evaluated the SCA values for the selected status of $\mathbf{\Lambda}$ over the testing data, and measured the WER values over the testing data using the selected DNN in the hybrid recognizer. 

In addition to experiments with our proposed training associated with the gLasso regularizer, we also tested (for comparison purposes) the training using the L2 regularizer for all of the weight matrixes and all of the bias vectors. 
%
%In addition to experiments with our proposed training associated with the gLasso regularizer, we also tested (for comparison purposes) the training using the L2 regularizer for the weight vectors of all hidden layers as well as $\mathbf{W}^{1}$ and $\{ \mathbf{b}^{l} \}^{l=1}_{L}$. 
%
%For the outgoing weight vector grouping, 
Then, the loss in (\ref{eq:regularizedLoss}) was replaced in this additional case by  
%
%\begin{equation}
%L( \mathbf{\Lambda} ) = E( \mathbf{\Lambda} ) + \sum_{l=1}^{L} \left\{ \alpha R_{\text{L2}}( \mathbf{W}^{l} ) + \beta R_{\text{L2}}( \mathbf{b}^{l} ) \right\} . 
%\label{eq:regularizedLossL2}
%\end{equation}
\vspace{-0.2cm}
\begin{equation}
L( \mathbf{\Lambda} ) = E( \mathbf{\Lambda} ) + \beta \sum_{l=1}^{L} \left\{ R_{\text{L2}}( \mathbf{W}^{l} ) + R_{\text{L2}}( \mathbf{b}^{l} ) \right\} . 
\label{eq:regularizedLossL2}
\end{equation}
\vspace{-0.2cm}

To elaborate the effects of our proposed training method, we also tested two different DNN sizes: one with 2048 nodes for every hidden layer ({\DNNt}) and another with 4096 nodes for every hidden layer ({\DNNf}). For both {\DNNt} and {\DNNf}, we evaluated the following four types of training: 1) with the gLasso regularizer for the outgoing hidden layer weight vectors ({\GLASSOo}), 2) with the gLasso regularizer for the incoming hidden layer weight vectors ({\GLASSOi}), 3) with the L2 regularizer for the outgoing hidden layer weight vectors ({\Lo}), and 4) with the L2 regularizer for the incoming hidden layer weight vectors ({\Li}). Particularly in the {\Lo} and {\Li} cases, we conducted common L2-regularized training (which corresponds to the training using (\ref{eq:regularizedLossL2})) but selected different nodes, based on the differences in grouping. Moreover, as stated in \ref{sec:DNNtraining}, we used in all four cases the L2 regularizer for all of the bias vectors; We also used the L2 regularizer for the input layer weights in the {\GLASSOo} case, and for the output layer weights in the {\GLASSOi} case.  

For convenience, we collectively refer to both the {\GLASSOo} and {\GLASSOi} training cases as gLasso and L2 for both the {\Lo} and {\Li} cases. 

%----------------%
\vspace{-0.1cm}
\subsection{Results}
\subsubsection{Fundamental performances}
%
%
%----------------%
\begin{table}[!t]
\renewcommand{\arraystretch}{1.0}
\caption{Achieved classification performances (\%) in {\DNNt}}
\label{tab:result}
\vspace{0.0mm}
\centering
\scalebox{1.0}[1.0]
{
\begin{tabular}{|c|c|c|c|c|}
\hline
{\scriptsize Regularizer}  & {\scriptsize \SCAb} & {\scriptsize \WERb} & {\scriptsize \SCAa} & {\scriptsize \WERa} \\
\hline
\hline
{\scriptsize \GLASSOo}	& {\scriptsize 46.2}  & {\scriptsize 18.7} & {\scriptsize 46.2} & {\scriptsize 18.8}  \\
\hline
{\scriptsize \GLASSOi}	&  {\scriptsize 46.1} & {\scriptsize 18.2} & {\scriptsize 46.1} & {\scriptsize 18.2}  \\
\hline
{\scriptsize \Lo}	& {\scriptsize 45.9}  & {\scriptsize 18.6} & {\scriptsize 22.8} & {\scriptsize 42.6}  \\
\hline
{\scriptsize \Li}	& {\scriptsize 45.9}  & {\scriptsize 18.6} & {\scriptsize 8.2} & {\scriptsize 100.0} \\ 
\hline
\end{tabular}
}
\vspace{-3.0mm}
\end{table}

%----------------%
%

In Table \ref{tab:result}, we summarize the SCA and WER values that we achieved for {\DNNt} in the four training cases: {\GLASSOo}, {\GLASSOi}, {\Lo}, and {\Li}. {\SCAb} and {\WERb} represent the SCA and WER values (\%), both obtained before the node selection; the SCA and WER values were obtained after the node selection for {\SCAa} and {\WERa}. 

First, we focus on the results in the {\SCAb} and {\WERb} columns. Regardless of the weight vector grouping type and the regularizer type, DNN training achieved such accurate classification performances as an SCA of about $46\%$ and a WER less than $19\%$.
They also show that the gLasso procedures achieved the performance that were competitive with the L2 procedures, which are widely used in the neural network training.

\begin{figure}[t]
\begin{minipage}[b]{1.0\linewidth}
  \centering
  \centerline{\includegraphics[width=5.5cm]{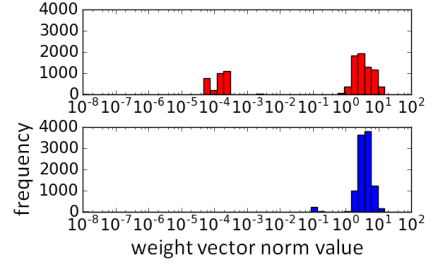}}
  \vspace{-3.0mm}
\end{minipage}
\caption{Histogram of outgoing weight vector norm values in {\GLASSOo} or {\Lo} cases. Training was done for {\DNNt}. Top: {\GLASSOo}, Bottom: {\Lo}.}
\label{fig:histogramOut}
\vspace{-5.0mm}
\end{figure}

Next, we observe the characteristics of the weight vector norm values that were obtained by the training. Figure \ref{fig:histogramOut} shows a histogram of the outgoing weight vector norm values in the {\GLASSOo} or {\Lo} cases for {\DNNt}.
The horizontal axis indicates the weight vector norm value, and the vertical axis indicates the frequency of its corresponding norm value. From this figure, we obtained the following findings: 
%\begin{enumerate}
1) The weight vector norm values by the {\GLASSOo} cases were clearly separated into two groups, suggesting the ease with which the threshold ($\theta$) can be set for the node selection between them;
2) The weight vector norm values reduced by the {\GLASSOo} procedure were rather small (at most slightly larger than $10^{-4}$). This distinct reduction proves that the {\GLASSOo} process successfully reduced the norm values of some selected weight vectors.
3) The weight vector norm values by the {\Lo} case were concentrated in the region between $10^{0}$ and $10^{1}$. This implies that the L2 procedure failed to reduce the weight vector norm values.
%\end{enumerate}

Due to space limitations, we omit the introduction of the results in the {\GLASSOi} and {\Li} cases. However, we obtained norm value distribution that closely resembled that in Figure \ref{fig:histogramOut}. 

%
%----------------%
%----------------%

We return to Table \ref{tab:result} and compare the results in the {\SCAb} and {\WERb} columns with those in the {\SCAa} and {\WERa} columns. 

We executed node selection by setting $\theta$ to $10^{-2}$, which is clearly far from the norm values in either of the two separate regions generated by the gLasso training. Using this threshold, we pruned 3161 nodes from the five hidden layers in the {\GLASSOo} case (the number of retained hidden layer nodes = 7079) and 3368 nodes in the {\GLASSOi} (number of retained hidden layer nodes = 6872) case.
%The performances, which improved after the node selection in the {\SCAa} and {\WERa} columns, are basically the same as those that improved before the node selection in the {\SCAb} and {\WERb} columns.
The performances gained after the node selection (in the {\SCAa} and {\WERa} columns) are basically the same with those gained before the node selection (in the {\SCAb} and {\WERb} columns).
The comparison here clearly proves that, regardless of the type of weight vector grouping, gLasso training sorted out the weight vectors by selectively reducing the norm values for some weight vectors and allowed node selection simply based on thresholding. 

As Figure \ref{fig:histogramOut} shows, pruning a certain amount of nodes is not easy from L2-based DNNs by setting a threshold. For comparison, we removed (in ascending order) the small-norm nodes from the DNN produced by the common L2-regularized training in the {\Lo} case until the number of removed nodes became identical to the {\GLASSOo} case: 3161; similarly, in the {\Li} case, we removed 3368 nodes from the L2-based DNN, which was the same with the network used in the {\Lo} case. The scores in the {\SCAa} and {\WERa} columns of Table \ref{tab:result} were obtained using these size-reduced L2-based DNNs. They are much worse than those gained before the node selection and clearly show that both the {\Lo} and {\Li} cases failed to sort out the weight vectors (as a result, their corresponding nodes).  

%\subsubsection{Characteristics of L2-based weight vectors}
%
%To further analyze the characteristics of the L2-based weight vectors, we observed the SCA scores along with changing the node selection threshold ($\theta$) for the {\Lo}-based {\DNNt}. Table \ref{tab:relationNodeDisposalAccuracy} shows the SCA scores obtained after the node disposal. In the table, we show the SCA score gained without the node disposal. The results clearly demonstrate that the SCA scores rapidly decrease when even a small number of nodes are disposed, and they suggest that the {\Lo}-based DNN represents salient information over most of the weight vectors in a widely dispersed manner. 
%
%----------------%
%\begin{table}[!t]
%\renewcommand{\arraystretch}{1.0}
%\caption{Relation between node disposal and classification power in {\Lo}-based {\DNNt}.}
%\label{tab:relationNodeDisposalAccuracy}
%\vspace{0.0mm}
%\centering
%\scalebox{1.0}[1.0]
%{
%\begin{tabular}{|c|c|c|c|c|c|}
%\hline
%$\theta$ & - & 1.0 & 2.0 & 3.0 & 4.0 \\
%\hline
%Num. disposed nodes	& 0 & 325  & 1241 & 1966 & 6010  \\
%\hline
%SCA	(\%) & 45.9 & 45.8 & 43.5 & 39.7 & 3.0  \\
%\hline
%\end{tabular}
%}
%\vspace{-0.0mm}
%\end{table}
%----------------%
%

\begin{figure}[t]
\begin{minipage}[b]{1.0\linewidth}
  \centering
  \centerline{\includegraphics[width=5.8cm]{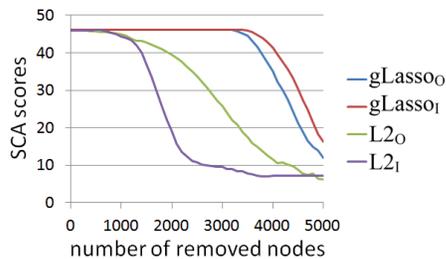}}
  \vspace{-3.0mm}
\end{minipage}
\caption{Effects of regularizer selection in node selection in {\DNNt}.}
\label{fig:pruningCurves} 
\vspace{-1.0mm}
\end{figure}

%----------------%
%\subsubsection{Effects of regularizer selection in reducing norm values}
%
To further analyze the effects of regularizer selection, we observed the changes in the SCA scores along with gradually removing hidden layer nodes, by 100 nodes, in increasing order of their norm values. 
Figure \ref{fig:pruningCurves} shows the SCA scores, each of which is a function of the number of removed nodes.
In the figure, the horizontal axis indicates the number of removed nodes; the SCA scores for the vertical axis. 
If the nodes do not exist in the small norm value region, the SCA scores gradually decrease as nodes are removed. On the other hand, if the nodes exist in the small norm value region, the SCA values remain high and later decrease. The curves in the figure clearly prove that the L2 procedure corresponds to the former situation and the gLasso procedure achieves the latter desirable situation. 

%----------------%
%----------------%
%

\subsubsection{Properties of gLasso procedure}
\begin{figure}[t]
\begin{minipage}[b]{1.0\linewidth}
  \centering
  \centerline{\includegraphics[width=5.5cm]{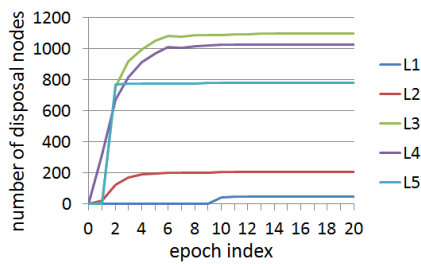}}
  \vspace{-3.0mm}
\end{minipage}
\caption{Number of disposal nodes for every hidden layer as a function of training epoch in {\GLASSOo}-based {\DNNt}.}
\label{fig:numDisposalNodesEpoch}
\vspace{-3.0mm}
\end{figure}

When introducing gLasso's training scheme in \ref{sec:DNNtraining}, we stated that the disposable nodes are disabled in some early training stage. To investigate this point, we observed the number of disposable hidden layer nodes for every hidden layer as a function of the training epoch. Here we define a disposable node as one whose norm value is smaller than $10^{-2}$. Figure \ref{fig:numDisposalNodesEpoch} illustrates the observed numbers for all five hidden layers (1st through 5th) of the {\GLASSOo}-based {\DNNt}. We found the following: 
1) Node selection was stably realized with the training progress.
2) It occurred in such comparatively early epochs as the 1st through 5th epochs.
3) It dominantly occurred in such higher hidden layers as the 3rd and 4th. 
%
%----------------%
%----------------%
%

\begin{figure}[t]
\begin{minipage}[b]{1.0\linewidth}
  \centering
  \centerline{\includegraphics[width=7.9cm]{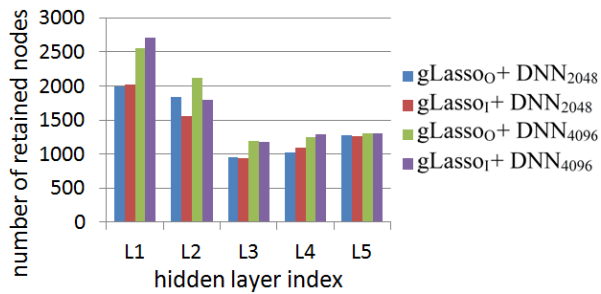}}
  \vspace{-3.0mm}
\end{minipage}
\caption{Number of finally retained hidden layer nodes in {\GLASSOo} and {\GLASSOi} cases.}
\label{fig:numSelecedNodesGLASSO}
\vspace{-3.0mm}
\end{figure}

The results in Figure \ref{fig:numDisposalNodesEpoch} suggest that the gLasso procedure shows a trend that mainly reduces the weight vector norm values in the higher hidden layers. To scrutinize this trend, in both the {\DNNt} and {\DNNf} cases, we pruned the disposable nodes using the $10^{-2}$ threshold and observed the number of selected (retained) nodes for every hidden layer. Figure \ref{fig:numSelecedNodesGLASSO} illustrates those numbers. Regardless of the weight vector grouping (outgoing vs. incoming) or the DNN size ({\DNNt} vs. {\DNNf}), we found that the gLasso procedure suppressed the weight vector norm values, specifically in the 3rd and 4th hidden layers, and achieved a size-reduced and reshaped network structure that was shared in all of the tested cases of {\GLASSOo}, {\GLASSOi}, {\DNNt}, and {\DNNf}. 
%
%----------------%
%----------------%

\vspace{-0.1cm}
\section{Related Work}
%
%A fundamental device for finding desirable DNN structures is to reduce the excess capability caused by too many parameters of DNN. 
%
An orthodox approach for the size reduction of a large matrix is to use Singular Value Decomposition. A large weight matrix in DNN is decomposed into small matrixes and re-expressed (in the sense of low-rank approximation) by replacing the small eigenvalues of the decomposed diagonal matrix by zero \cite{xue2013,xue2014}. This approach allows parameter reduction without serious degradation of the original discriminative power of the network. However, it does not reduce the number of hidden layer nodes, and therefore it is probably insufficient to achieve a desirable amount of network parameter reduction, especially in the case of using the network with the huge number of hidden layer nodes. 

For directly pruning the hidden layer nodes, the L1 norm of the weight vectors was used as an \textit{important} function. It resembles the weight-grouping-based norm that we use in our proposed training procedure \cite{tianxing2014}. However, because such mechanism of reducing the norms in the training procedure as ours was not implemented, DNN restructuring had to rely on additional retraining results. 

To find a desirable DNN structure, the Genetic Algorithm (GA) has been applied \cite{shinozaki2015}. It has high potential for generating a target structure, but it often needs an enormous amount of computational resources. 

Compared with the above related studies, our proposed training that embeds the gLasso regularizer is characterized by two main advantages: 1) Even if no additional training is conducted, it can easily produce node-pruned DNNs without performance degradation, and 2) its computation time is almost the same as conventional L2 regularization. 

Motivated by the gLasso procedure defined in linear regression \cite{yuan2006}, an approach similar to ours was proposed \cite{scardapane2016}. The gLasso regularizer was defined for the outgoing vector grouping of DNN and its effect was evaluated in various small- and middle-sized tasks that classify fixed-dimensional patterns. Compared to this work, we studied in this paper both the incoming and outgoing vector groupings and also elaborated our proposed scheme in a large-scale speech recognition environment. Here, incoming vector grouping can be considered a filtering function and will be appropriate for node selection or restructuring for Convolutional Neural Networks. 

%----------------%
\section{Conclusion}
\label{sec:conclusion}
We investigated the feasibility of the gLasso procedure, i.e., DNN training associated with the gLasso regularizer, for hidden layer weight vectors in a large-scale speech recognition task with the DNN-HMM hybrid speech recognizer in a TED Talks task. We elaborated the nature of gLasso training by applying it to both outgoing and incoming weight vector groupings and compared the gLasso and L2 procedures, where the L2 regularizer was used for all of the weight matrixes and all of the bias vectors. From the experiment results, we clearly demonstrated that our gLasso procedure automatically (without additional training) disabled less useful hidden layer nodes without degradation in DNN classification performance and successfully produced a small, necessary, and sufficient DNN structure. 

%\begin{center}
%\textbf{ACKNOWLEDGMENTS}
%\end{center}
%\vspace{-2.0mm}
%This work was supported in part by JSPS Grants-in-Aid for Scientific Research No. 26280063, MEXT-Supported Program
%``Driver-in-the-Loop'', and Grant-in-Aid for JSPS Fellows. The authors appreciate their financial support.

%\bibliographystyle{IEEEbib}
%\bibliography{refs}

\end{document}